\pdfoutput=1

\documentclass[11pt]{article}

\usepackage{EACL2023}

\usepackage{times}
\usepackage{latexsym}
\usepackage{graphicx}
\usepackage{amsfonts}
\usepackage{float}

\usepackage[T1]{fontenc}

\usepackage[utf8]{inputenc}

\usepackage{microtype}

\usepackage{inconsolata}
\usepackage{amsmath}

\title{The Shape of Learning: Anisotropy and Intrinsic Dimensions \\in Transformer-Based Models}

\author{
Anton Razzhigaev\textsuperscript{1,2},  
Matvey Mikhalchuk\textsuperscript{2,4}, 
Elizaveta Goncharova\textsuperscript{2,5}, \\
\bf Ivan Oseledets\textsuperscript{1,2}, 
\bf Denis Dimitrov\textsuperscript{2,3,4}, and
Andrey Kuznetsov\textsuperscript{2,3,6}\\
\textsuperscript{1}Skoltech,
\textsuperscript{2}AIRI, 
\textsuperscript{3}SberAI,\\
\textsuperscript{4}Lomonosov Moscow State University,\\
\textsuperscript{5}HSE University,\\
\textsuperscript{6}Samara National Research University\\
\href{mailto:razzhigaev@skol.tech}{razzhigaev@skol.tech} \\ 
}

\begin{document}
\maketitle
\begin{abstract}

In this study, we present an investigation into the anisotropy dynamics and intrinsic dimension of embeddings in transformer architectures, focusing on the dichotomy between encoders and decoders. Our findings reveal that anisotropy profile in transformer decoders exhibits a distinct bell-shaped curve, with the highest anisotropy concentrations in the middle layers. This pattern diverges from the more uniformly distributed anisotropy observed in encoders. In addition, we found that the intrinsic dimension of embeddings increases during the initial phases of training, indicating an expansion into the higher-dimensional space. Which is then followed by a compression phase towards the end of the training with dimensionality decrease, suggesting a refinement into more compact representations.  Our results provide fresh insights on the understanding of encoders and decoders embedding properties.\footnote{Accepted to EACL-2024}

\end{abstract}

\section{Introduction}

\begin{figure*}
    \centering
    \includegraphics[width=\linewidth]{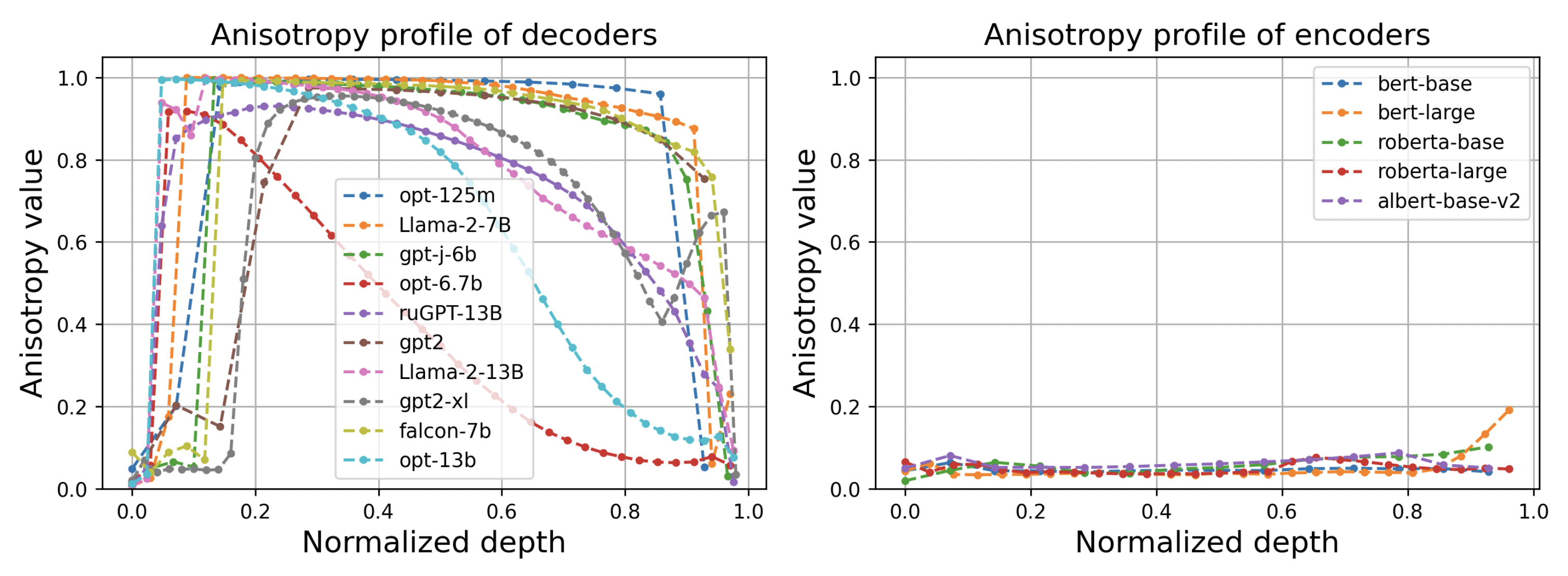}
    \caption{Different anisotropy profiles for transformer-based encoders and decoders.}
    \label{fig:encoder-decoders}
\end{figure*}

Introduced by \citet{attention_is_all_you_need}, the transformers have underpinned many breakthroughs, ranging from language modeling to text-to-image generation. As the adoption of transformers has grown, so has the pursuit to understand the intricacies of their internal mechanisms, particularly in the realm of embeddings.

Embeddings in transformers are intricate structures, encoding vast amounts of linguistic nuances and patterns. Historically, researchers have mainly examined embeddings for their linguistic capabilities \cite{semantic-probing,morphology-probing,syntactic-probing}. Yet, more nuanced properties lie beyond these traditional scopes, like anisotropy and intrinsic dimensionality, which can offer critical insights into the very nature and behavior of these embeddings.

Anisotropy, essentially representing the non-uniformity of a distribution in space, provides a lens, through which we can study orientation and concentration of the embeddings \cite{ethayarajh-2019-contextual,bis-etal-2021-much}. A higher degree of anisotropy suggests that vectors are more clustered or directed in specific orientations. In contrast, the intrinsic dimension offers a measure of the effective data dimensionality, highlighting the essence of information that is captured by the embeddings. Together, these metrics can serve as pivotal tools to probe into the black-box nature of transformers.

Our investigation uncovers the striking contrast in the anisotropy dynamics between transformer encoders and decoders. By analyzing the training phases of various transformer models, we shed light on the consistent yet previously unrecognized patterns of the anisotropy growth. Even more, our analysis reveals a unique dynamic of the averaged intrinsic dimension across layers in decoders: an initial growth during the early stages of training is followed by a decline towards the end. This suggests a two-phase learning strategy, where the model initially tries to unfold information in higher dimensional spaces and subsequently compresses it into more compact concepts, possibly leading to more refined representations.

\paragraph{Main Contributions:}
\begin{itemize}
    
    \item Uncovered a distinct bell-shaped curve for the anisotropy profile\footnote{Layer-wise anisotropy} in transformer decoders, contrasting with the uniformly distributed anisotropy in encoders.

    \item Confirmed that anisotropy increases progressively in the decoders as the training proceeds.
    
    \item Identified a two-phase dynamic in the intrinsic dimension of decoder embeddings: an initial expansion into higher-dimensional space, followed by a compression phase indicating a shift towards compact representations.

\end{itemize}

\section{Methodology}
\subsection{Datasets}


As our source for embedding we chose enwik8 dataset (English Wikipedia\footnote{\url{https://www.wikipedia.org/}}) that contains 100 million bytes of Wikipedia dump, making it a rich source of diverse textual content. It is publicly available through the Hutter Prize website\footnote{\url{http://prize.hutter1.net}}. The preprocessing stage includes the removal of all the code, media, and HTML tags, resulting in a clean and structured dataset with the vocabulary of 205 distinct characters.


\subsection{Embeddings}

The vectors are grouped into batches, each with a minimum of 4096 elements. We apply the selected method to determine anisotropy or intrinsic dimension to this batch. Prior to assessing intrinsic dimension, the embeddings are shuffled (before batching) to mitigate potential correlations. The results from individual batches are then averaged to calculate the metric for that layer, also capturing the standard deviation.

\subsection{Anisotropy}

To compute anisotropy, we employ the singular value decomposition (SVD).

Let $X \in \mathbb{R}^{n\_\mathrm{samples} \times \mathrm{emb}\_\mathrm{dim}}$ represent the centered matrix of embeddings, where $\sigma_1, \dots, \sigma_k$ are its singular values. The anisotropy score of $X$ is given by:

\[
\text{anisotropy}(X) = \frac{\sigma_1^2}{\sum_{i=1}^{k} \sigma_i^2}.
\]

Equivalently, this can be deduced using the eigenvalues $\sigma_1^2, \dots, \sigma_k^2$ of the covariance matrix:
\[
C = \frac{X^T X}{n\_\mathrm{samples} - 1}.
\]

For some models, we compare the anisotropy measurement approach based on the SVD decomposition with the average cosine \cite{ethayarajh-2019-contextual,bis-etal-2021-much} between embeddings for each layer.
\[
\text{average\_cosine} = \frac{2}{n(n-1)}\sum_{1\leq i<j\leq n}\cos(X_i, X_j),
\]
where  $X_i$ and $X_j$ denote two vectors of embeddings of the same layer (these vectors can originate from different contexts and correspond to different model inputs).

We also study the effect of the centering (subtraction of average vector from embeddings before calculations) for these two types of metrics.

\subsection{Intrinsic Dimension}

To determine the intrinsic dimension of a set of embeddings, we utilize the approach proposed by \citet{DBLP:journals/corr/abs-1803-06992}. This method explores how the volume of an $n$-dimensional sphere (representing the count of embeddings) scales with dimension \( d \).

For each data point within our embeddings, we determine the distances \( r_1 \) and \( r_2 \) to their two closest neighboring points. This process generates a set of pairs \( \{(r_1, r_2)\} \). Using this set, the intrinsic dimension \( d \) can be estimated. Firstly, we define:
\[
\mu_i = \frac{r_2}{r_1},
\]
for each point \( i \).

The cumulative distribution function (CDF) of \( \{\mu_i\} \) is provided by:
\[
F(\mu) = (1 - \mu^{-d}) \mathbf{1}_{[1,+\infty)}(\mu).
\]
This expression for \( F \) is based on the derivations and proofs presented by the authors of the referenced paper. From the CDF, we deduce:
\[
\frac{\log(1 - F(\mu))}{\log(\mu)} = d.
\]

To estimate \( d \), linear regression \( y=kx \) is applied on the plane \( (x, y) \), with:
\[
x_i = \log(\mu_i) \quad \text{and} \quad y_i = 1 - F_{\text{emp}}(\mu_i),
\]
where \( F_{\text{emp}} \) signifies the empirical CDF for \( \{\mu_i\} \).

For some models, we also measure the intrinsic dimension by other local methods. We use Manifold-adaptive dimension estimation \cite{farahmand_2007} and Method of Moments \cite{amsaleg_2018}.

All three local methods show correlating results in our experiments.

\section{Related Work}

\subsection{Isotropy of Hidden Representations}

\citet{gao2019representation} introduce the \textit{representation degeneration problem}. This is the phenomenon of degenerating in the representation of learned embeddings in the generative models, particularly when they are tied. The authors conclude that, unlike fixed word embeddings (e.g., word2vec \cite{mikolov2013efficient}), vanilla transformer embeddings are clustered within the narrow cone.

Recent research revealed that global anisotropy is a common trait among all transformer-based architectures \cite{ait-saada-nadif-2023-anisotropy,godey2023anisotropy, multimodal-isotropy}. However, within the local subspaces, isotropy prevails, enhancing model expressiveness and contributing to high performance in the downstream tasks.

\citet{ding-etal-2022-isotropy} conducted an extensive empirical evaluation of modern anisotropy calibration methods, showing no statistically significant improvements in the downstream tasks. They conclude that the local isotropy of the hidden space of transformers may lead to the high level of model's expressiveness \cite{cai2021isotropy}. While most isotropy findings are observed in encoder-only or encoder-decoder architectures, \citet{cai2021isotropy} brought an interesting variation to light. The authors conducted experiments on various architectures, evaluating the reduced effective embedding dimension using PCA, and observed high cosine values across the layers, especially in models such as GPT-2 (decoder). 

The work \citep{ait-saada-nadif-2023-anisotropy} supports previous research through extensive experimental evaluation. This study arose from the presence of local isotropy in hidden representations, suggesting that anisotropy does not necessarily compromise the expressiveness of these representations.

\citet{godey2023anisotropy} investigated the potential causes of anisotropy, particularly its connection to rare words in the model's vocabulary. They explored character-level models to eliminate the influence of rare tokens, but these models did not show any significant improvements in the experiments. The authors also uncovered that adding common bias term to the inputs can lead to the increased attention score variance, promoting the emergence of categorical patterns in self-attention softmax distributions. Increasing input embeddings norm shows signs of anisotropy based on the query and key values.

\subsection{Intrinsic Dimensionality}
Following the idea of local isotropy of the hidden representations, the investigation of the intrinsic task-specific subspaces offers new insights into the fine-tuning and also the potential to improve model efficiency. \citet{Chunyuan2018} suggested that the training trajectory of Transformer architectures occurs in a low-dimensional subspace. \citet{zhang-etal-2023-fine} demonstrated that fine-tuning engages only a small portion of the model’s parameters, and it is possible to identify the principal directions of these intrinsic task-specific subspaces. Using their method of identifying the training direction they achieved performance similar to the fine-tuning in the full parameter space. 

\citet{tulchinskii2023intrinsic} used intrinsic dimension estimation to identify AI-generated texts. Specifically, they utilized the persistent homology dimension estimator \cite{schweinhart2019persistent} as the tool for assessing dimensionality. The findings revealed that the intrinsic dimension of natural texts tends to cluster between higher values in comparison to generated texts. The latter exhibits a lower dimension, irrespective of the specific generator involved.

\subsection{Training Progress}

Prior research has utilized information criteria to investigate the internal regularization mechanisms of neural networks. \citet{shwartzziv2017opening} delve into simple fully connected networks and advocate for identifying a trade-off between information compression and prediction at each layer of the network. They claimed that a significant portion of training epochs in deep fully-connected networks focuses on compressing the input into an efficient representation rather than fitting the training labels.

In \cite{achille2019critical}, the authors found that the training process of deep neural networks is not monotonic when it comes to information memorization. They identified two distinct stages in the training process. The initial stage is marked by rapid information growth, resembling a memorization procedure, while the subsequent stage involves a reduction of information — referred to as ``reorganization'' or ``forgetting'' by the authors.

This findings is on par with our observations regarding the two-phase training of the language models, where the intrinsic dimension experiences initial growth followed by a subsequent decline. Notably, during this phase, the model's performance exhibits steady improvement (see Section \ref{section:ID} and Figure~\ref{fig:ID_methods}).

\subsection{Encoder and Decoder Architectures}

The original transformer architecture consists of both encoder and decoder blocks, and each of these blocks can operate independently. The self-attention mechanism is a shared key feature, with decoders utilizing causal self-attention. Decoders are typically trained for language modeling tasks, focusing on generating coherent sequences of the text. In contrast, encoders are aimed to produce contextual representations (i.e., embeddings), from the input text.


Taking limited previous research on the distinctions between the inner representations of encoders and decoders into account, our study analyzes multiple encoder-based models (such as BERT \cite{devlin-etal-2019-bert}, RoBERTa \cite{liu2019roberta}, and ALBERT \cite{lan2020albert}), and decoder-based models (including OPT 125M-13B \cite{zhang2022opt}, Llama-2 7B-13B, Llama-2 7B Chat \cite{touvron2023llama}, GPT2 \cite{radford2019language}, GPT-J \cite{gpt-j}, Falcon-7B, and Falcon-7B-Instruct \cite{falcon7b}) to offer a comprehensive comparison of their behavior.

\begin{table*}[ht!]
    \centering
    \small

\resizebox{0.8\textwidth}{!}{
\small{
    \begin{tabular}{|l|c|c|c|c|c|c|} \hline 
 & Bloom-560M & Bloom-1.1B & Bloom-3B & Bloom-7B & Pythia-2.8B & TinyLlama-1.1B\\
 \hline
 \multicolumn{7}{c}{\textit{Architecture hyperparameters}}\\
 \hline
 Layers & 24 & 24 & 30 & 30 & 32 &  22 \\\hline
 Hidden dim. & 1024 & 1536 & 2560 & 4096 & 2560 &  2048 \\\hline
 Attention heads & 16 & 16 & 32 & 32& 32 &  16 \\\hline
 Activation & \multicolumn{4}{c|}{GELU} & GELU & SwiGLU \\\hline
 Vocab size & \multicolumn{4}{c|}{250,680} & 50,257 & 32,000 \\\hline
 Context length & \multicolumn{4}{c|}{2048} & 2048 & 2048 \\\hline
 Position emb. & \multicolumn{4}{c|}{Alibi} & RoPE & RoPE \\\hline
 Tied emb. & \multicolumn{4}{c|}{True} & False & False \\
 \hline
 \multicolumn{7}{c}{\textit{Pretraining hyperparameters}}\\
 \hline
 Global Batch Size & 256 & 256 & 512 & 512 & 1024 &  1024 \\\hline
 Learning rate & 3.0e-4 & 2.5e-4 & 1.6e-4 & 1.2e-4 & 1.6e-4 &  4.0e-4 \\\hline
 Total tokens & \multicolumn{4}{c|}{341B} & 300B & 3T \\\hline
 Warmup tokens & \multicolumn{4}{c|}{375M} & 3B & 4B \\\hline
 Min. learning rate & \multicolumn{4}{c|}{1.0e-5} & 1.6e-5 &  4.0e-5 \\
 \hline
 \end{tabular}}
}
\caption{Architectural and training configurations of the analyzed models.}
\label{tab:parameters}
\end{table*}

\section{Results}

\begin{figure}
    \centering
    \includegraphics[width=\linewidth]{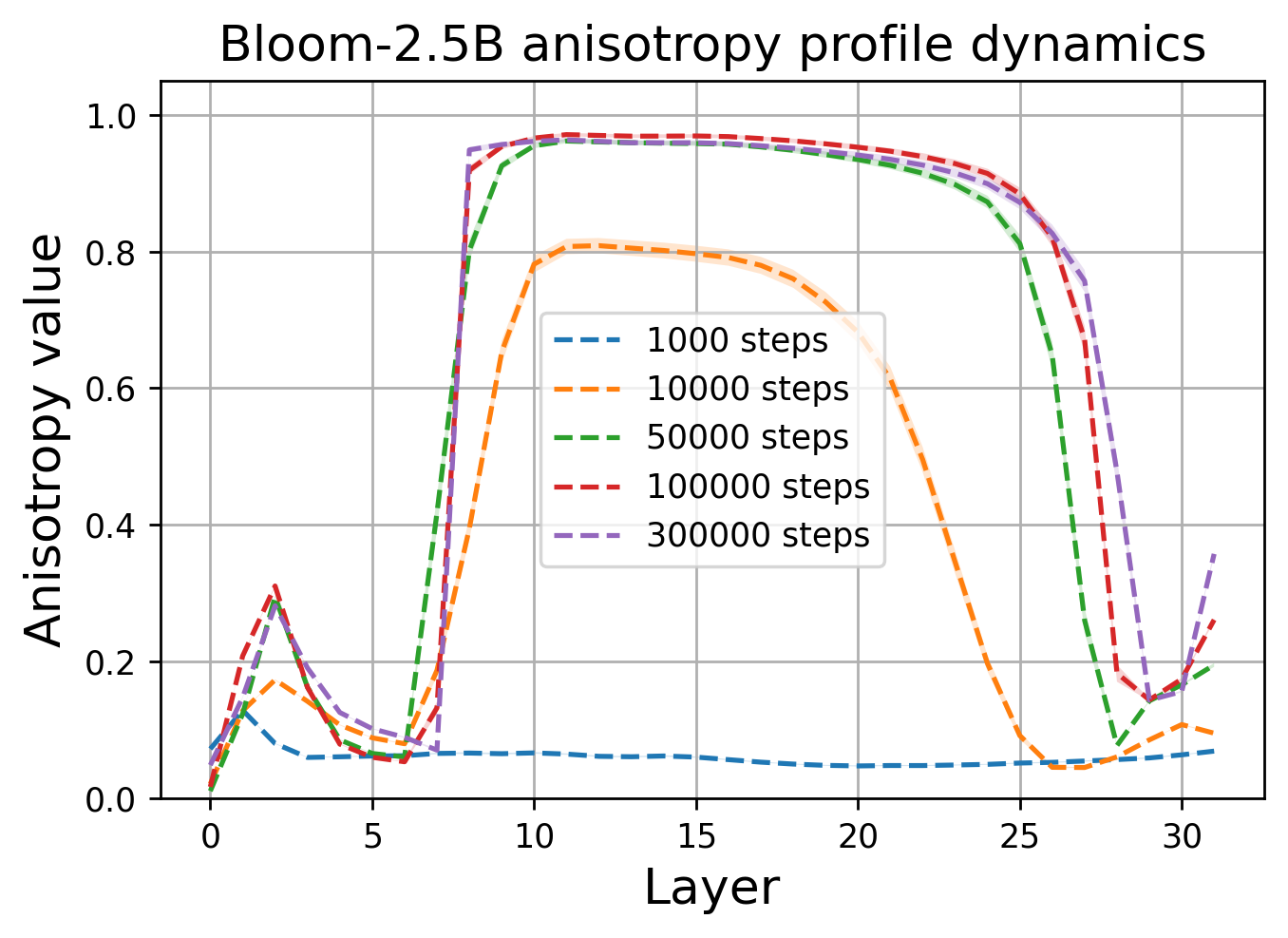}
    \caption{Anisotropy profile for Bloom-3B at different number of pretraining steps.}
    \label{fig:Bloom-anisitropy}
\end{figure}

\begin{figure}
    \centering
    \includegraphics[width=\linewidth]{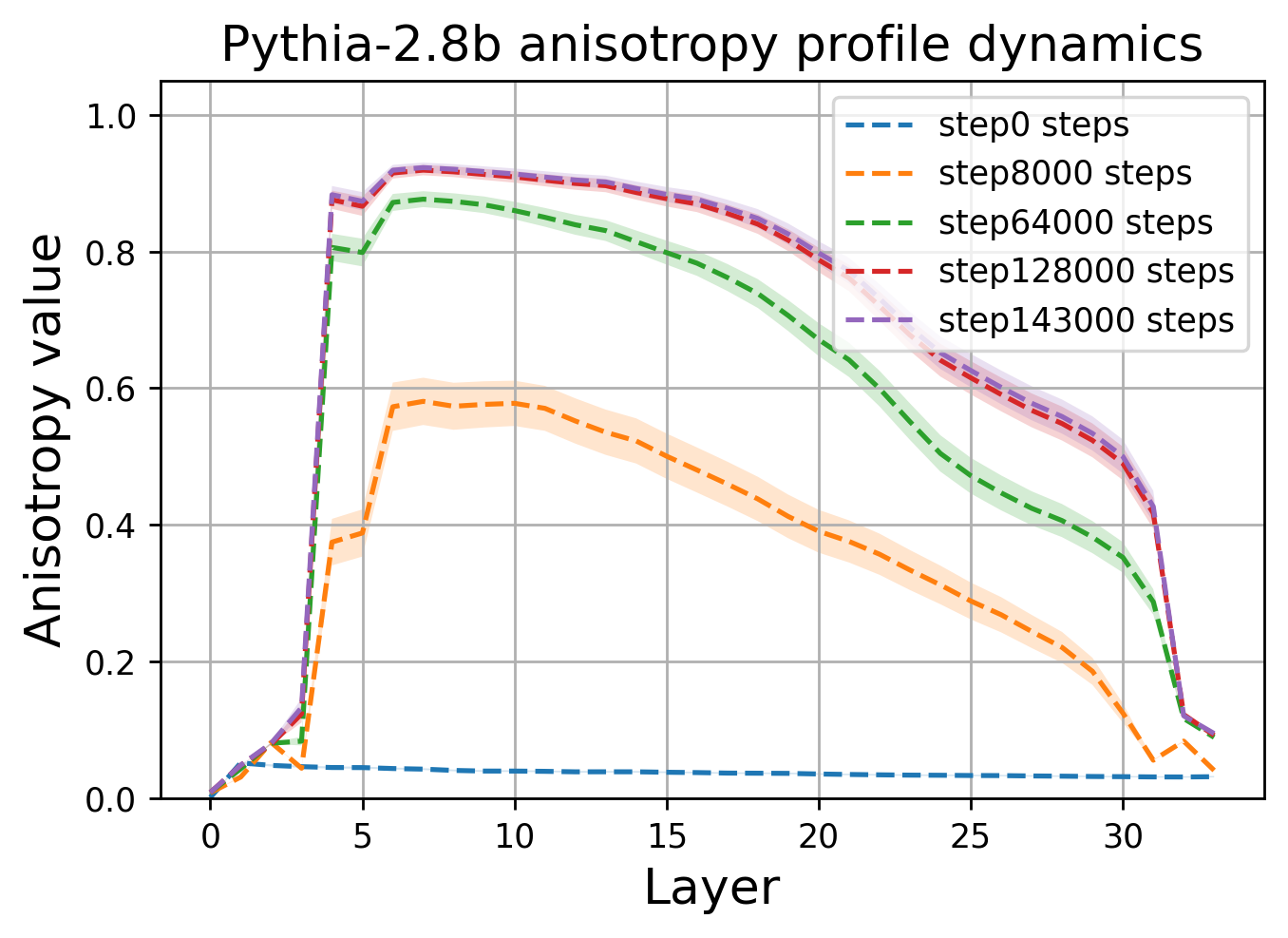}
    \caption{Anisotropy profile for Pythia-2.8B at different number of pretraining steps.}
    \label{fig:pythia-anisitropy}
\end{figure}

\begin{figure}
    \centering
    \includegraphics[width=\linewidth]{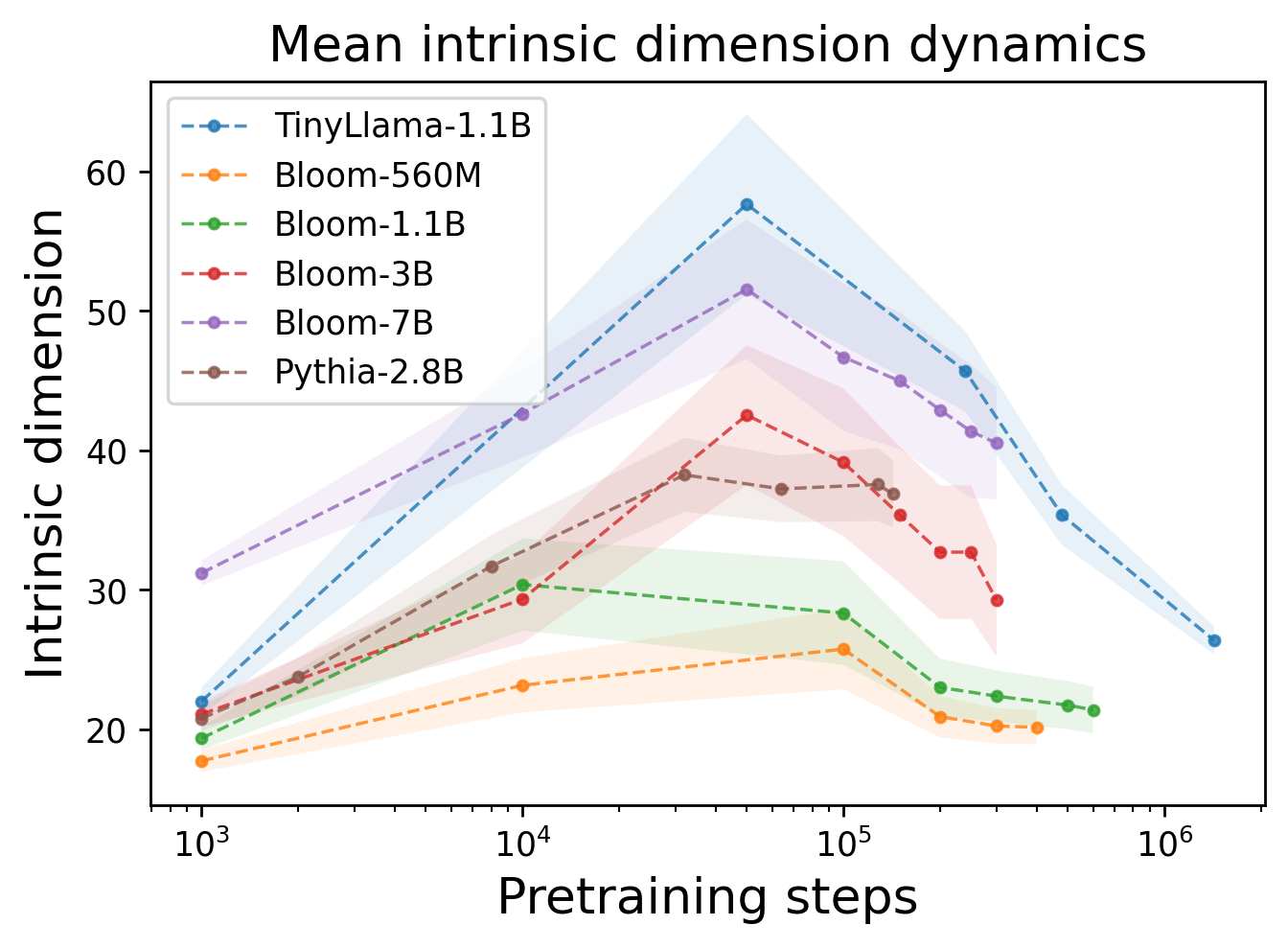}
    \caption{Intrinsic dimension averaged across layers at different pretraining steps.}
    \label{fig:ID}
\end{figure}

In this section, we present our empirical findings concerning the anisotropy dynamics and intrinsic dimensionality of transformer embeddings at different layers. Our results span various pretrained transformer models, showcasing clear patterns in the behavior of encoders versus decoders, and illuminating the transformation of their properties during the training process.

\subsection{Anisotropy Across Pretrained Transformers}

We began by comparing the anisotropy levels across various pretrained transformers, analyzing both encoder and decoder models. Their anisotropy profiles can be found in the Figure~\ref{fig:encoder-decoders}.

\textbf{Encoders}: Anisotropy levels remain relatively consistent across the models, with minor variations based on the model size and training data.
    
\textbf{Decoders}: In contrast to the encoders, decoders showcase a unique bell-shaped structure, indicating that the middle layers tend to have a higher anisotropy concentration among all examined models.

\subsection{Anisotropy Dynamics During Training}

To further probe the evolution of anisotropy, we examine its progression through the training phases of various models.

Figure~\ref{fig:Bloom-anisitropy} and Figure~\ref{fig:pythia-anisitropy} capture this trajectory by plotting anisotropy values for decoders at different training checkpoints at all internal layers. The consistent growth pattern, followed by stabilization, is observed across various models, suggesting an inherent characteristic of the language modeling training dynamics of decoders.



\subsection{Intrinsic Dimensionality During Training}
\label{section:ID}

Our exploration into the intrinsic dimensionality reveals intriguing patterns: Figure~\ref{fig:ID} displays the averaged intrinsic dimension of models throughout the training process. The initial stages exhibit a sharp rise, indicating the model's attempt to map the information to higher dimensional spaces. However, as training progresses, there is a notable decline, suggesting a subsequent phase where the model compresses this information, refining more compact concepts.

\subsection{Model Architecture}

For the conducted research, we analyze decoder-based models with similar parameter scales but different architectural and training configurations. In Table~\ref{tab:parameters}, we summarize the main solutions for the models presented in Figure~\ref{fig:ID}.

It is noteworthy that there is a considerable difference among models with the same number of parameters (Bloom-1.1B and TinyLlama-1.1B), each featuring distinct architectural configurations. The intrinsic dimension of the latter is higher both at the end of training and at its peak. The obtained results also lead to the conclusion that the growth and the decline of the intrinsic dimension do not show correlation with the warmup period in the learning rate scheduler.

\section{Conclusion}

Our exploration into the anisotropy dynamics and intrinsic dimensionality of transformer embeddings has brought significant distinctions between encoder and decoder transformers to light. Notably, the intrinsic dimensionality showcases a two-phased training behaviour, where models initially expand information into higher-dimensional spaces and then refine it into compact concepts towards the end of training. These insights not only deepen our understanding of transformer architectures but also opens up new avenues for tailoring training approaches in future NLP research.

\section*{Limitations}

While our study offers valuable insights into the behavior of transformer embeddings, there are a few limitations to consider.

\textbf{Model Diversity}: Our findings predominantly revolve around specific transformer models, and generalization to all transformer architectures is not guaranteed.
    
\textbf{Training Dynamics}: The observed two-phased behavior in intrinsic dimensionality might be influenced by the datasets or specific training configurations.
    
\textbf{Anisotropy Interpretation}: While we identified distinct anisotropy patterns in encoders and decoders, the direct implications of these patterns on downstream tasks remain to be fully explored.

\section*{Ethics Statement}

Our research focuses on analyzing transformer embeddings and does not involve human subjects or sensitive data. All findings are derived from publicly available models and datasets. We strive for transparency and reproducibility in our methods and analyses.

\bibliography{anthology,custom}
\bibliographystyle{acl_natbib}

\appendix
\section{Alternative ID and Anisotropy Estimation Methods}

\begin{figure}[H]
    \centering
    \includegraphics[width=.9\linewidth]{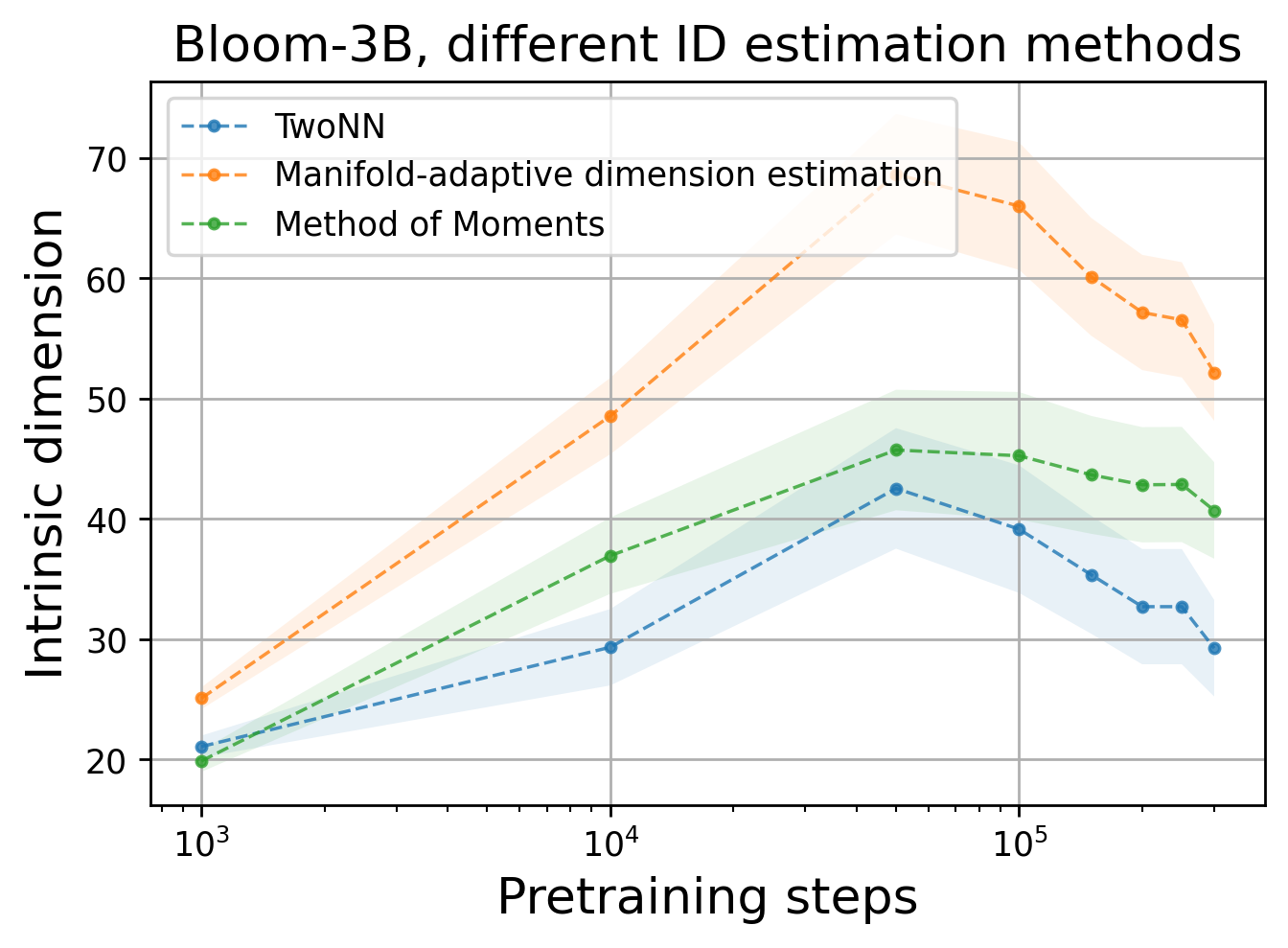}
    \caption{Intrinsic dimension (ID) averages across layers at different pretraining steps estimated via 3 different algorithms.}
    \label{fig:ID_methods}
\end{figure}

\end{document}